\documentclass{article}
\usepackage{spconf,amsmath,graphicx}

\usepackage{algorithmic}
\usepackage{algorithm}
\usepackage{booktabs}
\usepackage{amsfonts}

\title{Promoting Cooperation in Multi-Agent Reinforcement Learning via Mutual Help}

\name{Yunbo Qiu\qquad Yue Jin\qquad Lebin Yu\qquad Jian Wang\qquad Xudong Zhang}
\address{Department of Electronic Engineering, Tsinghua University, Beijing, China}
\begin{document}

\maketitle

\begin{abstract}

Multi-agent reinforcement learning (MARL) has achieved great progress in cooperative tasks in recent years. 
However, in the local reward scheme, where only local rewards for each agent are given without global rewards shared by all the agents, traditional MARL algorithms lack sufficient consideration of agents' mutual influence. In cooperative tasks, agents' mutual influence is especially important since agents are supposed to coordinate to achieve better performance.
In this paper, we propose a novel algorithm Mutual-Help-based MARL (MH-MARL) to instruct agents to help each other in order to promote cooperation.
MH-MARL utilizes an expected action module to generate expected other agents' actions for each particular agent. Then, the expected actions are delivered to other agents for selective imitation during training.
Experimental results show that MH-MARL improves the performance of MARL both in success rate and cumulative reward.

\end{abstract}


\section{Introduction}

In recent years, multi-agent reinforcement learning (MARL) has achieved great success in cooperative tasks, such as electronic games\cite{vinyals2019grandmaster}, power distribution\cite{wang2021multi}, and cooperative navigation\cite{jin2019efficient}.
The design of the reward scheme is critical in MARL to instruct agents to fulfill corresponding objectives.
Depending on the specific application, local reward scheme\cite{lowe2017multi,ackermann2019reducing} or global reward scheme\cite{foerster2018counterfactual,rashid2018qmix} is designed to provide rewards for individual agents or rewards shared by all the agents, respectively. The global reward scheme involves the problem of credit assignment \cite{foerster2018counterfactual}.
Traditional algorithms with the local reward scheme optimize agents' local rewards which lack explicit reflection on the mutual influence between agents, and thus result in insufficient coordination among agents\cite{sheikh2020multi}.
Mutual influence of agents is essential in MARL, especially in cooperative tasks, where agents are expected to coordinate to achieve better performance from the perspective of the whole multi-agent system.

There are mainly three approaches that incorporate explicit mutual influence among agents into traditional MARL: communication methods, teacher-student methods, and local-global reward-based methods. Communication methods improve the cooperation of agents through communication protocols between agents.
\cite{sukhbaatar2016learning} trains agents' policies alongside the communication.
\cite{jiang2018learning} further studies on when to communicate and how to use communication messages.
\cite{kim2020communication} forms each agent's intention and then communicates it to other agents.
However, communication methods require explicit communication channels to exchange messages between agents, which may not be available in many applications.

Teacher-student methods instruct agents to learn from each other. \cite{torrey2013teaching} guides agents in learning from agents who are already experienced. \cite{da2017simultaneously, omidshafiei2019learning} consider the framework where agents teach each other while being trained in a shared environment. \cite{zhu2020learning,ilhan2021action} conduct deeper research on how to reuse received instructions from teacher agents.
In \cite{christianos2020shared}, agents implicitly teach each other via sharing experiences.
However, teacher-student methods require agents to be identical to directly utilize other agents' instructions on actions. 

Local-global reward-based methods simultaneously utilize local rewards and global rewards which can be constructed by local rewards. \cite{wang2022individual} learns two policies with local rewards and global rewards respectively, and restricts the discrepancy between the two policies. 
\cite{sheikh2020multi} maintains a centralized critic to learn from global rewards and a decentralized critic to learn from local rewards, and simultaneously optimizes the two critics.
However, local-global reward-based methods require the design of global rewards based on the local reward scheme. In addition, they only consider the relationship between a single agent and the whole multi-agent system from the perspective of the reward scheme, without detailed relationships among agents.

In this paper, we propose a novel algorithm Mutual-Help-based MARL (MH-MARL) to instruct agents to help each other while being trained by traditional MARL algorithms, where an expected action module is added to traditional MARL algorithms. In traditional MARL with the local reward scheme, agents only optimize their own policies, ignoring other agents' performance. In contrast, each agent learns to generate actions that can help increase other agents' returns while optimizing its own return in MH-MARL. Specifically, each agent selectively imitates the actions that are expected by other agents while being trained by a traditional MARL algorithm. The expected actions are generated by an expected action module. The selective imitation follows the principle that an agent chooses to learn to help others if helping others does not seriously harm its own performance.
Note that MH-MARL considers mutual help between agents without the requirement of communication during execution. Besides, agents are not required to be identical.
We evaluate MH-MARL in a flocking navigation environment\cite{iot} where cooperation between agents is crucial for success. Experimental results demonstrate that MH-MARL improves performance in both success rate and reward.
The main contributions of this paper are listed as follows:
\begin{itemize}
\item A novel algorithm MH-MARL is proposed to promote cooperation in the local reward scheme of MARL by utilizing an expected action module to instruct agents to selectively help each other while optimizing their own policies.
\item Experiments are conducted to verify the effectiveness of MH-MARL in the cooperation task and its capability to be built on different fundamental MARL algorithms.

\end{itemize}

\section{Background}
\subsection{Markov Games with Local Reward Scheme}
Markov Games with the local reward scheme is generally modeled similarly to \cite{lowe2017multi, ackermann2019reducing}. At each time step $t$, each agent $i$ chooses action $a_i$ to interact with the environment based on its local observation $o_i$ and agent's own policy $\pi_i(o_i)$. 
The combinations of each agent's observation $o_i$ and each agent's action $a_i$ are denoted as joint observation $\boldsymbol{o}$ and joint action $\boldsymbol{a}$, respectively.
After agents' interaction with the environment, each agent $i$ receives a new local observation $o'_i$ and a local reward $r_i$.
The goal of each agent's policy $\pi_i$ is to maximize the expectation of its return, which is also known as cumulative reward: $R_i=\sum_{t=0}^\infty \gamma^t r_{i,t}$, where $\gamma$ is a discount factor.

\subsection{Multi-Agent Reinforcement Learning Framework}

Actor-critic architecture and centralized training and decentralized execution (CTDE) are two commonly used components in MARL framework.

In actor-critic architecture, each agent $i$ learns an ‘Actor’ and a ‘Critic’. The ‘Actor’ represents a policy function $\pi_i$ that can generate action $a_i$ according to observation $o_i$. The ‘Critic’ represents an action-value function, also known as Q-function, which estimates the return of a policy given current actions $\boldsymbol{a}$ and observations $\boldsymbol{o}$.

In CTDE\cite{lowe2017multi}, during training, agents have access to other agents' observations and actions. This helps agents to estimate the expected return more accurately from a more comprehensive perspective.
During execution, each agent is required to choose its action merely based on its own observation, which enables agents to act in a distributed manner.

MADDPG\cite{lowe2017multi} is a representative MARL algorithm with the local reward scheme.
The Q-functions in MADDPG are optimized by minimizing the following loss:
\begin{equation}
\begin{aligned}
L_{i, critic}&=\mathbb{E}_{\boldsymbol{o},\boldsymbol{a},r_i,\boldsymbol{o}'\sim
\mathcal{D}}[(Q_i (\boldsymbol{o},a_1,...,a_n)-y_{i})^2],\\
\end{aligned}
\label{equ-critic-on}
\end{equation}
where $\mathcal{D}$ is a replay buffer to store and reuse experiences of agents, and $y_{i}$ is the target value obtained by:
\begin{equation}
\begin{aligned}
y_{i}&=r_i+\gamma Q'_i (\boldsymbol{o}',a'_1,...,a'_n)|_{a'_j=\pi'_j(o'_j)},\\
\end{aligned}
\label{equ-y-on}
\end{equation}
where $Q'_i$ and $\pi'_j$ are target functions as backups to stabilize training.

The policy functions in MADDPG is optimized by minimizing the following loss:
\begin{equation}
\begin{aligned}
L_{i, actor}&=\mathbb{E}_{\boldsymbol{o},\boldsymbol{a}\sim \mathcal{D}}[-Q_i(\boldsymbol{o},a_1,...,a_i,...,a_n)|_{a_i=\pi_i(o_i)}].\\
\end{aligned}
\label{equ-actor-on}
\end{equation}

\section{Algorithm}

Traditional MARL algorithms do not consider active and explicit mutual influence of agents' actions on each other sufficiently, although Q-functions take other agents' actions and observations into account in CTDE mainly for the problem of non-stationarity. Agents' mutual influence is essential, especially in cooperative tasks, where agents are supposed to achieve better performance through teamwork.
In the local reward scheme, where agents only optimize their own returns, the lack of consideration of mutual influence is more obvious. 

\begin{figure}[t]
\vspace{-0.1in}
\centerline{\includegraphics[width=0.35\textwidth]{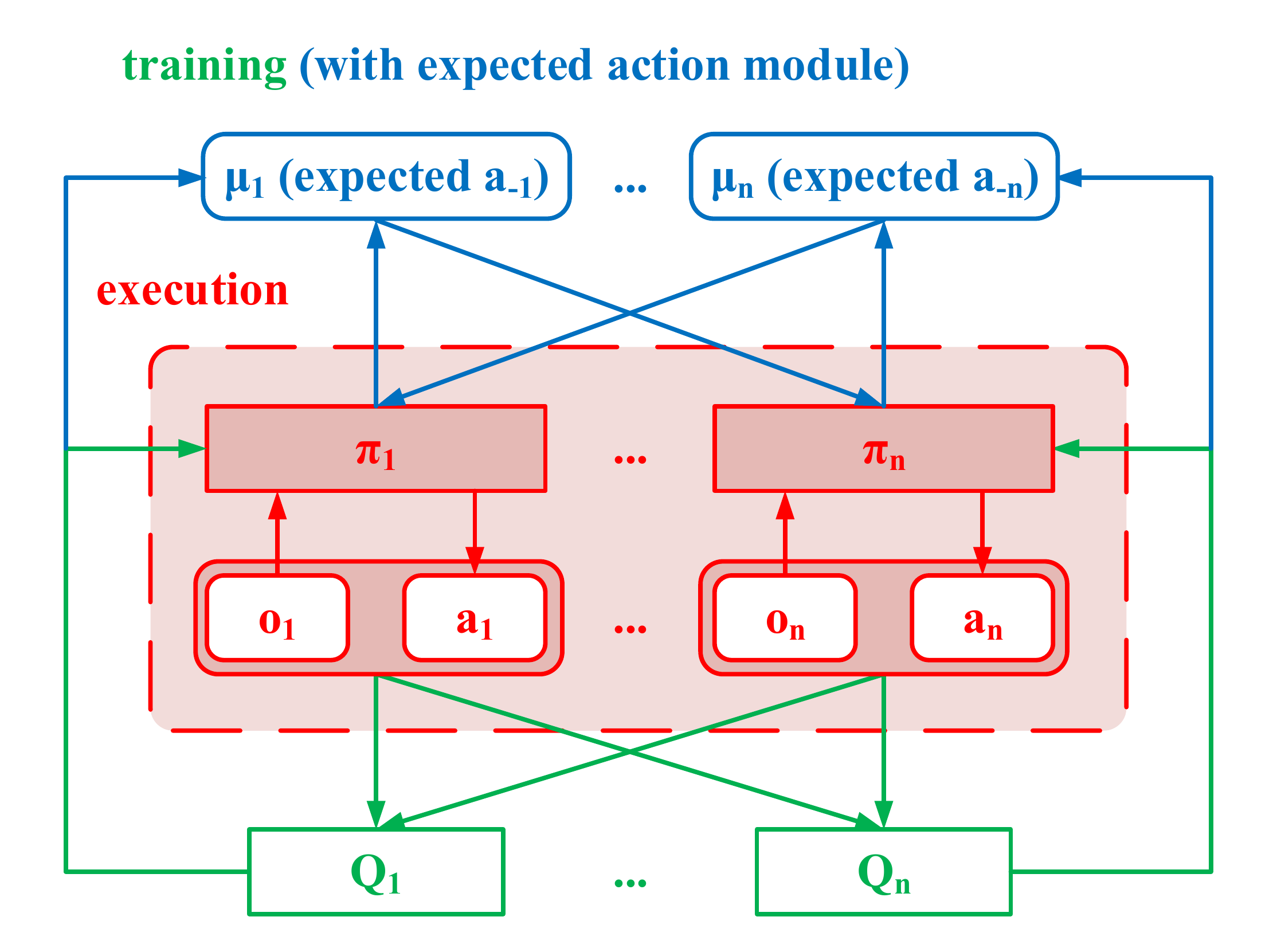}}
\vspace{-0.1in}
\caption{Overview of algorithm MH-MARL.
Actor module and critic module in traditional MARL are colored in red and green, respectively. Blue color represents expected action module in MH-MARL which generates expected actions for others to selectively imitate.
}
\label{diag}
\vspace{-0.1in}
\end{figure}

Therefore, we propose a novel algorithm Mutual-Help-based MARL (MH-MARL) to instruct agents to help each other by adding an expected action module to traditional actor-critic architecture, as shown in Fig.~\ref{diag}. MH-MARL aims to guide each agent in helping others without severely harming its own performance, and thus achieves the effect of ‘one for all, all for one’ in cooperative tasks. MH-MARL can be built on any traditional MARL algorithms with actor-critic architecture and CTDE in the local reward scheme. In this paper, we choose to build MH-MARL on a representative MARL algorithm MADDPG\cite{lowe2017multi}. Mutual help with expected action module during training consists of two processes: generating the expected actions for each agent and utilizing generated expected actions from other agents.

Firstly, we focus on how to generate the expected actions. 
The Q-function of each agent involves all the agents' actions as a part of input in CTDE, which represents that other agents' actions have an impact on the return of the agent. Thus, if each agent expects to maximize its own performance, the agent can rely on the adjustment of other agents' actions, apart from the adaption of the agent's own actions.
To make it clear which actions of other agents can help an agent to the utmost extent, each agent $i$ maintains an expected policy function $\mu_i$ to generate expected actions of other agents according to joint observation $\boldsymbol{o}$ by minimizing the following loss:
\begin{equation}
\begin{aligned}
L_{i, expect}&=\mathbb{E}_{\boldsymbol{o}\sim \mathcal{D}}[-Q_i(\boldsymbol{o},a_i,\boldsymbol{a_{-i}})|_{a_i=\pi_i(o_i),\boldsymbol{a_{-i}}=\mu_i(\boldsymbol{o})}],\\
\end{aligned}
\label{expect}
\end{equation}
where experiences are randomly extracted from the replay buffer $\mathcal{D}$ as (\ref{equ-actor-on}), $a_i$ is generated by the agent's current policy $\pi_i$, and $\boldsymbol{a_{-i}}$ represents actions of other agents expected by agent $i$. As the progress of training, the Q-functions will be more accurate as trained by minimizing (\ref{equ-critic-on}), and then $\mu_i$ can generate the expected actions of other agents more accurately.

Secondly, we consider how to utilize expected actions from other agents when each agent optimizes its own policy function $\pi_i$. For each agent $i$, if we consider optimizing $\pi_i$ only to help another agent $j$, then the action generated by $\pi_i$ should imitate $\mu_j(\boldsymbol{o},i)$ (expected action for the agent $i$ in $\mu_j(\boldsymbol{o})$) , which is realized by minimizing the discrepancy between $\mu_j(\boldsymbol{o},i)$ and agent's own action $\pi_i(o_i)$.
However, $\mu_j(\boldsymbol{o},i)$ may severely harm the performance of agent $i$ in some cases, and thus full adoption of expected actions will frustrate the optimization of agent $i$.
Therefore, the agent $i$ only considers imitating the expected actions which do not severely decrease its own performance. This selective imitation is realized by minimizing the following loss:
\begin{equation}
\begin{aligned}
L_{i,help}=&\mathbb{E}_{\boldsymbol{o},\boldsymbol{a}\sim \mathcal{D}}[\frac{1}{n-1} \sum_{j \neq i }[\left | \left |  \pi_{i}(o_{i}) - \mu_{j} (\boldsymbol{o}, i)  \right | \right |_{2}\\
&\cdot \epsilon(Q_i(\boldsymbol{o},a_1,...,a_i,...,a_n)|_{a_i=\mu_j(\boldsymbol{o},i)}+\eta \\
&- Q_i(\boldsymbol{o},a_1,...,a_i,...,a_n)|_{a_i=\pi_i(o_i)})]],\\
\end{aligned}
\label{others}
\end{equation}
where $\epsilon(x)=max(0,\frac{x}{|x|})$ is a step function for comparison, and $\eta$ is a positive constant. Since each agent $i$ can receive expected actions from all the other agents, the selective imitation is averaged over all these expected actions.

In addition to helping others, each agent also optimizes its policy function to maximize its own return by minimizing (\ref{equ-actor-on}) as traditional MARL. The selective imitation to help others and the traditional MARL to optimize its own return are fulfilled simultaneously using the following loss function:
\begin{equation}
\begin{aligned}
L_{i,actor}^{MH}&=L_{i,actor}+\alpha_{i}\cdot L_{i,help},\\
\end{aligned}
\label{equ-actoe-whole}
\end{equation}
where $\alpha_{i}$ is the mix ratio of the two loss terms. To automatically balance the two loss terms, $\alpha_{i}$ is calculated every minibatch to convert the two loss terms to the same scale:
\begin{equation}
\begin{aligned}
\alpha_{i}&=\frac{\beta}{M}\sum_{\boldsymbol{o},\boldsymbol{a} \sim \mathcal{D}}|Q_i(\boldsymbol{o},a_1,...,a_i,...,a_n)|_{a_i=\pi_i(o_i)}|,\\
\end{aligned}
\end{equation}
where $M$ is the size of the minibatch and $\beta$ is a hyperparameter.
The whole algorithm of MH-MARL built on MADDPG is shown in Algorithm \ref{MH-MARL}.

\begin{algorithm} [t]
\footnotesize
	\caption{MH-MADDPG (MH-MARL built on MADDPG)} 
	\label{MH-MARL} 
	\begin{algorithmic}
	    \STATE \textbf{Initialize} parameters $\theta_i$ of each agent $i$
	    \FOR {episode $= 1 \text{ to } E$ }
	    \STATE Initialize a random process $\mathcal{N}$ for action exploration
        \STATE Receive initial observation $\boldsymbol{o}$
        \FOR{$t$ = 1 to $T_{episode}$}
        \STATE For each agent $i$, select action $a_i = \pi_{i}(o_i) + \mathcal{N}_t$ w.r.t. the current policy and exploration
        \STATE Execute actions $\boldsymbol{a} = (a_1, . . . , a_n )$ and obtain reward $\boldsymbol{r}$ and new observation $\boldsymbol{o}'$
        \STATE Store $(\boldsymbol{o}, \boldsymbol{a}, \boldsymbol{r}, \boldsymbol{o}')$ in replay buffer $\mathcal{D}$
        \STATE $\boldsymbol{o} \leftarrow \boldsymbol{o}'$
	    \FOR {agent $i = 1 \text{ to } n$ }
	    \STATE Randomly sample a mini-batch $M$ from $\mathcal{D}$
	    \STATE Generate expected actions for agent $i$ with every $\mu_{j \neq i}(\boldsymbol{o}, i)$
	    \STATE Update actor by minimizing (\ref{equ-actoe-whole}) with expected actions
        \STATE Update critic by minimizing (\ref{equ-critic-on}) 
        \STATE Update expected action policy by minimizing (\ref{expect})
	    \ENDFOR
	    \STATE Softly update target networks for each agent $i$: \\
	    $\theta '_i \leftarrow \tau \theta _i + (1-\tau) \theta '_i$
	    \ENDFOR
	    \ENDFOR
	\end{algorithmic} 
\end{algorithm}

\section{Experiments}

\subsection{Main Experiments}

To verify the effectiveness of MH-MARL, we conduct experiments in a flocking navigation environment\cite{iot}, where agents are required to navigate to a target area while maintaining their flock without collisions. It is a highly cooperative task since agents need to precisely control the distances between them to avoid both collisions and destroying the flock.
We build our MH-MARL on a representative MARL algorithm MADDPG\cite{lowe2017multi} to form MH-MADDPG, and contrast it with four baseline algorithms: MADDPG, MADDPG-GR, DE-MADDPG, and SEAC-MADDPG. 
MADDPG-GR uses a global reward which is the sum of local rewards and shared by all agents.
DE-MADDPG\cite{sheikh2020multi} simultaneously learns decentralized Q-functions to optimize local rewards and centralized Q-functions to optimize global rewards (which is the sum of local rewards as MADDPG-GR).
SEAC-MADDPG is designed based on MADDPG to directly share experiences between agents\cite{christianos2020shared}.
$\eta=0.05$ and $\beta=2$ as hyperparameters. 
All the algorithms are run in 5 seeds.

\begin{figure}[t]
\vspace{-0.1in}
\centerline{\includegraphics[width=0.45\textwidth]{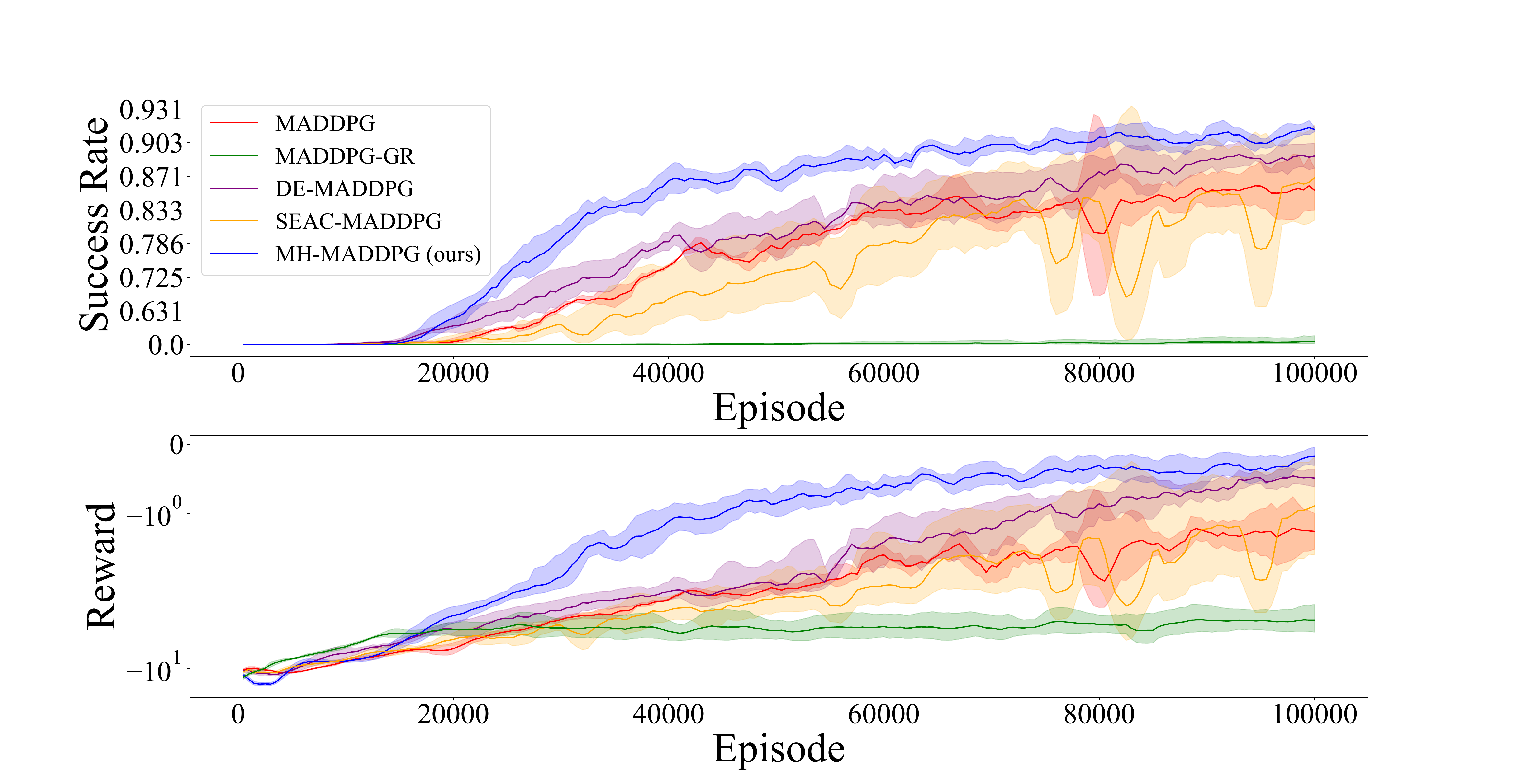}}
\vspace{-0.1in}
\caption{Convergence curves of success rate and reward of main experiments, presented in the fifth root scale and the symmetric log scale, respectively.}
\label{main}
\vspace{-0.1in}
\end{figure}

Convergence curves of success rate and reward are plotted in Fig.~\ref{main}. 
It shows that our algorithm MH-MADDPG learns the fastest during training and achieves the best success rate and reward after training, in comparison with four baseline algorithms. Directly transforming the local reward scheme to a global reward scheme (MADDPG-GR) results in poor performance. Both DE-MADDPG and SEAC-MADDPG surpass their fundamental MARL algorithm MADDPG in success rate and reward, but are still inferior to MH-MADDPG. Besides, due to the utilization of other agents' experience, the stability of SEAC-MADDPG is the worst. 

\subsection{Ablation Experiments}

To validate the effectiveness of all the parts of MH-MARL, we design three ablation algorithms of MH-MADDPG: ‘no help’ refers to the algorithm removing the expected action module in MH-MARL, which is exactly MADDPG; ‘no selectivity’ removes the step function in (\ref{others}) for full adoption of expected actions; ‘no MARL’ removes the traditional MARL loss term (\ref{equ-actor-on}) in (\ref{equ-actoe-whole}) to only help others.

\begin{figure}[t]
\vspace{-0.1in}
\centerline{\includegraphics[width=0.45\textwidth]{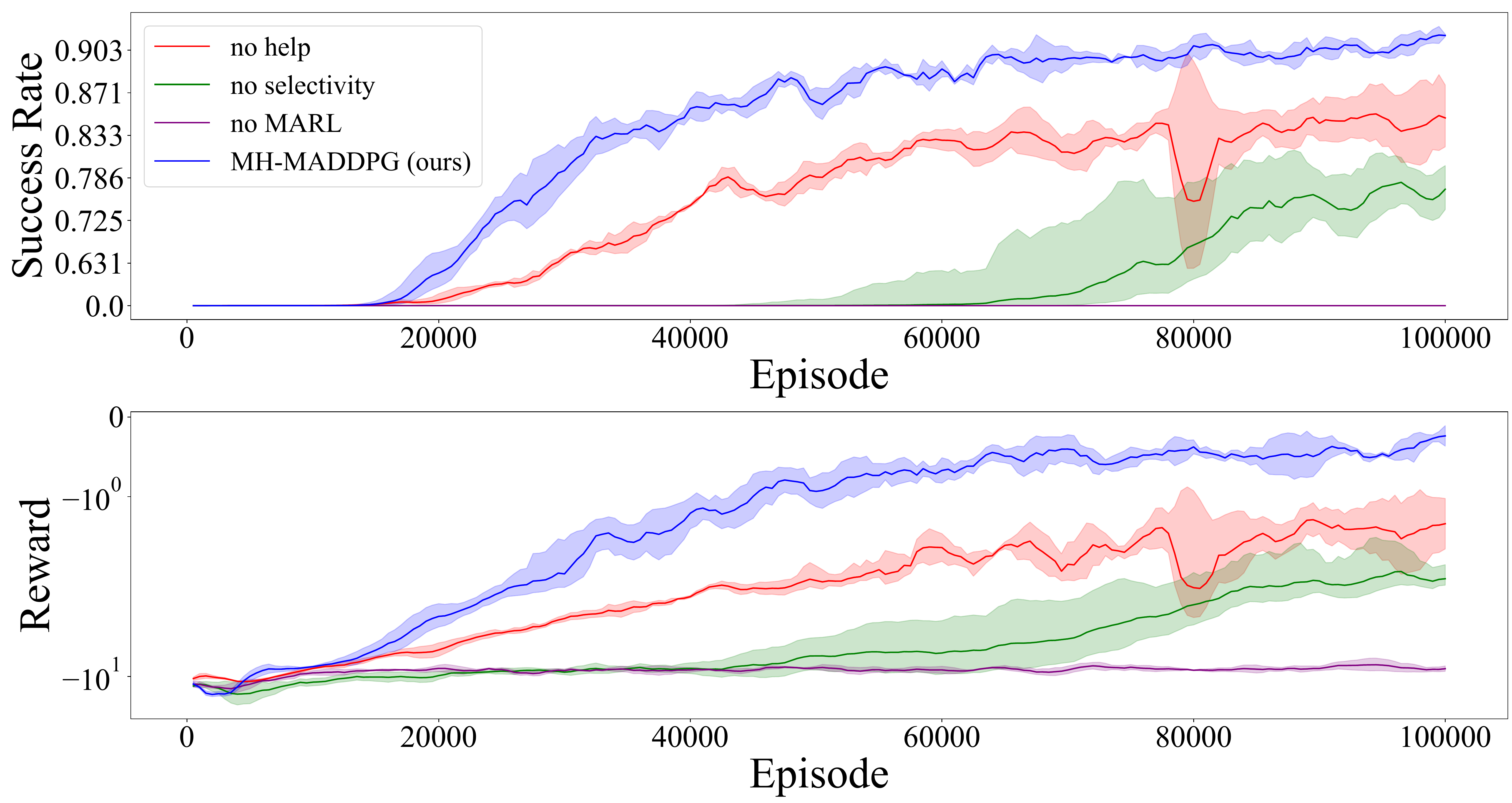}}
\vspace{-0.1in}
\caption{Convergence curves of success rate and reward of ablation experiments.}
\label{abla}
\vspace{-0.0in}
\end{figure}

Convergence curves of ablation experiments are plotted in Fig.~\ref{abla}. It demonstrates that MH-MADDPG is largely superior to three ablation algorithms throughout the training process, which validates the effectiveness of all the parts of MH-MARL. Specifically, in contrast to ‘no help’ (MADDPG), ‘no MARL’ can hardly improve the performance by training, and ‘no selectivity’ hinders the training in the early stage. It shows the importance of emphasis on the agent's own performance compared to helping others.

\subsection{Experiments with another fundamental MARL algorithm}

In previous experiments, we choose MADDPG as the fundamental MARL algorithm to implement our MH-MARL. However, MH-MARL is designed to be able to build on any MARL algorithm with actor-critic architecture and CTDE in the local reward scheme. 
In this subsection, we choose to build MH-MARL on MATD3\cite{ackermann2019reducing}, instead of MADDPG. Baseline algorithms are also adjusted to use MATD3 as their fundamental MARL algorithms.

\begin{figure}[t]
\vspace{-0.1in}
\centerline{\includegraphics[width=0.45\textwidth]{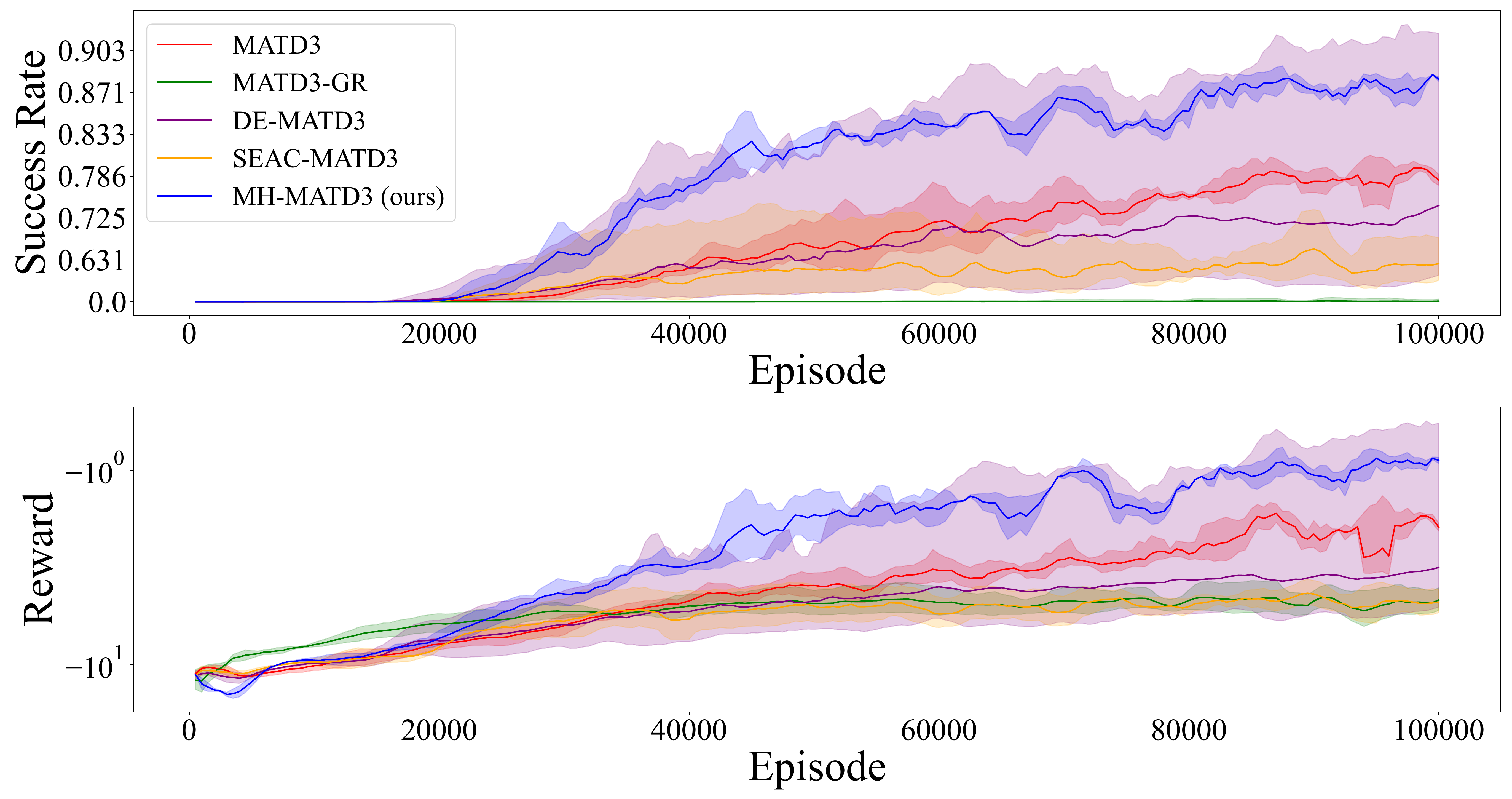}}
\vspace{-0.1in}
\caption{Convergence curves of success rate and reward of experiments with MATD3 as the fundamental MARL algorithm.}
\label{matd3}
\vspace{-0.1in}
\end{figure}

Convergence curves of success rate and reward are plotted in Fig.~\ref{matd3}. The figure demonstrates that MH-MATD3 still surpasses the baseline algorithms by a large margin in success rate and reward, which validates the flexibility in the choice of the fundamental MARL algorithm.
Specifically, DE-MATD3 and SEAC-MATD3 are even inferior to their fundamental algorithm MATD3, and the stability of DE-MATD3 is poor, unlike MH-MATD3.

\section{Conclusion and Future Work}

We propose a novel algorithm MH-MARL, which utilizes an expected action module to instruct agents to help each other. Agents first generate expected actions for other agents, and then other agents selectively imitate corresponding expected actions while being trained by a traditional MARL algorithm. MH-MARL emphasizes optimizing an agent's own performance as the primary goal, and mutual help is regarded as the secondary goal to promote cooperation.
Experiments are conducted to verify the superiority of MH-MARL in the improvement of performance in the cooperative task. Experimental results also show that MH-MARL can be built on different fundamental MARL algorithms.
In the near future, we expect to apply MH-MARL to other cooperative tasks.

\vfill\pagebreak

\bibliographystyle{IEEEbib}
\bibliography{ref}

\end{document}